\documentclass{article}

\usepackage[final,nonatbib]{nips_2017}

\usepackage{resizegather}
\usepackage{graphicx}
\usepackage[english]{babel}
\usepackage[utf8x]{inputenc}
\usepackage[T1]{fontenc}
\usepackage{fullpage}
\usepackage{bbold}
\def\decide#1#2{
  \mathrel{
    \mathop{
      \begin{array}{c}
        >\vspace{-1.4ex}\\<
      \end{array}
      }\limits_{#2}\limits^{#1}
    }
  }

\usepackage{amsmath,amssymb}
\usepackage{amsopn}
\usepackage{bbm}
\usepackage{amsthm}
\usepackage{graphicx}
\usepackage{subfigure}
\usepackage{multirow}

\usepackage[usenames,dvipsnames]{xcolor}
\usepackage[colorinlistoftodos]{todonotes}
\usepackage{algorithm}
\usepackage{algorithmic}

\newtheorem{theorem}{Theorem}
\newtheorem{lemma}[theorem]{Lemma}

\title{Dynamic Model Selection for Prediction Under a Budget}

\author{
  Feng Nan \\
  Systems Engineering\\
  Boston University\\
  Boston, MA 02215\\
  \texttt{fnan@bu.edu} \\
	\And
	Venkatesh Saligrama \\
	Electrical Engineering\\
	Boston University \\
	Boston, MA 02215 \\
	\texttt{srv@bu.edu} \\
}

\begin{document}

\maketitle

\begin{abstract}
We present a dynamic model selection approach for resource-constrained prediction. 
Given an input instance at test-time, a gating function identifies a prediction model for the input among a collection of models. Our objective is to minimize overall average cost without sacrificing accuracy. We learn gating and prediction models on fully labeled training data by means of a bottom-up strategy. Our novel bottom-up method is a recursive scheme whereby a high-accuracy complex model is first trained. Then a low-complexity gating and prediction model are subsequently learnt to {\it adaptively approximate} the high-accuracy model in regions where low-cost models are capable of making highly accurate predictions. We pose an empirical loss minimization problem with cost constraints to jointly train gating and prediction models. On a number of benchmark datasets our method outperforms state-of-the-art achieving higher accuracy for the same cost.
\end{abstract}

\section{Introduction}
Resource costs arises during test-time prediction in a number of machine learning applications. 
Feature costs in Internet, Healthcare, and Surveillance applications arise due to 
to feature extraction time~\cite{xu2013cost}, and 
feature/sensor acquisition~\cite{trapeznikov:2013b}. 
Data analytics applications in mobile devices are often performed on remote cloud services~\cite{phoneanal} due to the limited device capabilities. While these remote services have substantial resources for building highly complex and accurate predictive models, utilizing them is expensive.


One approach to offset these costs is to train stand-alone low-cost models capable of predicting under a budget constraint, but such models often have unacceptably poor accuracy.
%
%
Among the most promising approaches are adaptive methods that dynamically allocate resources based on the difficulty of the input instance to ensure high prediction accuracy with low overall cost~\cite{Gao+Koller:NIPS11,DBLP:conf/icml/XuWC12,trapeznikov:2013b,Wang2014,ASTC_AAAI14,NanNIPS2016}. 


Motivated by this perspective we develop a dynamic model selection approach for resource constrained prediction. At test-time, for each input instance, a {\it low-cost} gating function identifies a prediction model among a given collection of models ranging from low-cost, low-accuracy to high-cost, high-accuracy predictors to make a prediction. 
Since both prediction and gating account for cost, to reduce duplications, we favor design strategies that lead to shared features and decision architectures between the gating function and low-cost prediction models.

We learn gating and prediction models by training on fully annotated training datasets. We pose the problem as a discriminative empirical risk minimization problem that jointly optimizes for gating and prediction models. We propose a novel bottom-up recursive scheme 
whereby a high-accuracy complex model is first trained. Then low-complexity gating and prediction models are subsequently learnt so as to \emph{adaptively} approximate an existing model by identifying regions of input space where a low-complexity gating and prediction model are likely to result in high-accuracy. 

We formulate a joint margin-based optimization objective that is separately convex in gating and prediction functions. We propose an alternating minimization scheme that is guaranteed to converge since with appropriate choice of loss-functions (for instance, logistic loss), each optimization step amounts to a probabilistic approximation/projection (I-projection/M-projection) onto a probability space. While our method can be recursively applied in multiple stages to successively refine cost-accuracy trade-off we observe that on benchmark datasets even a single stage of our method outperforms state-of-art in cost-accuracy performance.

\noindent
{\it Bottom-up \& Top-Down:}
As described a bottom-up approach seeks to approximate a high-accuracy system with a lower cost model by determining regions of input space where a low-cost model suffices. In contrast the prevalent top-down approach~\cite{Gao+Koller:NIPS11,DBLP:conf/icml/XuWC12,ASTC_AAAI14} builds out from a low-cost (and low-accuracy) model and seeks to improve accuracy by selectively adding features. While both are viable approaches, bottom-up methods are particularly well-suited for reducing costs in the high-accuracy regime. 
Additionally, in many applications where we have proven high-accuracy legacy systems, a bottom-up method is natural to employ to reduce costs without sacrificing accuracy. 



%

\section{Related Work}
Learning decision rules to minimize error subject to a budget constraint during prediction-time is an area of active interest
\cite{Gao+Koller:NIPS11,DBLP:conf/icml/XuWC12,trapeznikov:2013b,weiss_taskar2013,Wang2014, NIPS2015_5982,ASTC_AAAI14,NanNIPS2016}.

{\it Pre-trained Models:} In one instantiation of these methods it is assumed that there exist a collection of prediction models with amortized costs~\cite{weiss_taskar2013,trapeznikov:2013b} so that a natural ordering of prediction models can be imposed. In other instances, the feature dimension is assumed to be sufficiently low so as to admit an exhaustive enumeration of all the combinatorial possibilities ~\cite{Wang2014,NIPS2015_5982}. These methods then learn a policy to choose amongst the ordered prediction models. In contrast we do not impose any of these restrictions. 

\noindent
{\it Top-Down Methods:}
For high-dimensional spaces, many existing approaches focus on learning complex adaptive decision functions top-down \cite{Gao+Koller:NIPS11,DBLP:conf/icml/XuWC12,ASTC_AAAI14,NIPS2015_5982}. Conceptually, during training top-down methods acquire new features based on their utility value. This requires exploration of low-cost feature subsets and sequentially add features that improve accuracy. This stage-wise exploration of feature subsets is often a combinatorial problem requiring relaxations and greedy heuristics. 

{\it Bottom-up Methods:}
A bottom-up approach is presented in \cite{NanNIPS2016} based on pruning random forests (RF) to reduce costs. Nevertheless, their perspective is to compress the original model and utilize the pruned forest as a stand-alone model for test-time prediction. 
In contrast our goal is to adaptively approximate the existing high-cost high-accuracy model and selectively utilize it in conjunction with our low cost model to achieve improved performance. This is particularly important for applications with pre-existing high-accuracy legacy systems.

The teacher-student framework \cite{LopSchBotVap16} is also related to our bottom-up approach; a low-cost student model learns to approximate the teacher model so as to meet test-time budget. However, the goal there is to learn a better stand-alone student model. In contrast, we make use of both the low-cost (student) and high-accuracy (teacher) model during prediction via a gating function, which learns the limitation of the low-cost (student) model and consult the high-accuracy (teacher) model if necessary, thereby avoiding accuracy loss.

\section{Problem Setup}
We consider the standard learning scenario of
resource constrained prediction. We assume that the training and test points are drawn i.i.d. according to some fixed but unknown distribution
${\cal D}$ over ${\cal X}\times {\cal Y}$. 
Given an instance $x \in {\cal X}$, at test-time, we have the option of selecting which model, among a collection of varying cost and accuracy models, to utilize to make a prediction. 
The prediction accuracy of a prediction model $f$ is modeled by a loss function $\ell(f(x),y)$. We utilize fully labeled training data to train our prediction and gating models by minimizing the empirical loss. 

Let us consider a single stage of our training method in order to formalize our setup. Consider a gating likelihood function $q(z|x)$ with $z \in \{0,\,1\}$, that outputs the likelihood of sending the input $x \in {\cal X}$ to  a prediction model, $f_z$.  
The model, $f_0$, is a complex model, which is either a priori known, or which we train to high-accuracy regardless of cost considerations. Our task is reduced to training gating, $q$, and the prediction model, $f_1$, to adaptively approximate $f_0$ so that the overall loss does not increase while the cost is reduced. Let $\ell(f_z(x),y)$ where $z \in \{0,1\}$ be the loss suffered on $x \in {\cal X}$ upon using the model $f_z(\cdot)$ and let $c(f_z,q,x)$ be the corresponding prediction cost, including the cost of using the gating likelihood function. The overall empirical loss can be written as:
\begin{align*}
\mathbb{E}_{S_n} \mathbb{E}_{q(z | x)} [\ell(f_z(x),y)] =\mathbb{E}_{S_n} [\ell(f_0(x),y)]+ \mathbb{E}_{S_n} \big[q(1 | x) \underbrace{(\ell(f_1(x),y) - \ell(f_0(x),y))\big]}_{\textcolor{red}{Excess\, Loss}}
\end{align*}
The first term only depends on $f_0$, and from our perspective a constant. 
Similar to average loss we can write the average cost as
\begin{align*}
\mathbb{E}_{S_n}\mathbb{E}_{q(z|x)} [c(f_z,q,x)] = \mathbb{E}_{S_n} [c(f_0,q,x)]-\mathbb{E}_{S_n} [q(1|x)\underbrace{(c(f_0,q,x)-c(f_1,q,x))}_{\textcolor{red}{Cost\, Reduction}}]
\end{align*}
If the gating cost is a negligible fraction of prediction cost, the first term is a constant. We can characterize the optimal gating function (see \cite{trapeznikov:2013b}) that minimizes the overall average loss subject to average cost constraint:
$$
\overbrace{\ell(f_1,x)-\ell(f_0,x)}^{\textcolor{red}{Excess\, loss}} \decide{q(1|x)=0}{q(1|x)=1} \eta \overbrace{(c(f_0,q,x)-c(f_1,q,x))}^{\textcolor{red}{Cost\, reduction}} 
$$
for a suitable choice $\eta \in \mathbb{R}$. This characterization encodes the important principle that if the marginal cost reduction is smaller than the excess loss, we prefer the high-accuracy predictor. 
Nevertheless, this characterization is generally infeasible. Note that both the LHS and RHS depends on knowing how well the high-accuracy model, $f_0$, performs on the input instance. Since this information is unavailable this target can be unreachable with low-cost gating. 

Rather than directly enforcing a low-cost structure on $q$ we decouple the constraint and introduce a parameterized family of gating functions $g(x) \in {\cal G}$ that attempts to mimic (or approximate) $q$. 

We can now pose our general objective as an optimization problem. We let $f_1 \in {\cal F}$ denote the family of low-cost prediction models. 
\vspace{-0.25cm}
\begin{equation}
\hspace{-.5cm}\begin{array}{rlll}\tag{OPT1}\label{eq:OPT1}
\displaystyle \min_{f_1 \in {\cal F}, g \in {\cal G}, q}& \mathbb{E}_{S_n} \sum\limits_{z}[ q(z|x)\ell(f_z(x),y)] + \overbrace{D(q(\cdot |x),g(x))}^{\textcolor{red}{gating\, approx}} \\ 
\textrm{subject to: } & \mathbb{E}_{S_n}\mathbb{E}_{q(z|x)}[ c(f_z,g,x)] \leq B.\,\,\,   \textcolor{red}{(budget)}
\end{array}
\end{equation}
The objective function penalizes excess loss and ensures through the second term that this excess loss can be enforced through admissible gating functions. 
The constraint requires that the learnt gating and prediction models must satisfy the required budget constraint.
In many cases the cost $c(f_z,g,x)$ is independent of the example $x$ and depends primarily on the model being used. This is true for linear models where each $x$ must be processed through the same collection of features. For these cases $c(f_z,g,x) \triangleq c(f_z,g)$. The budget constraint simplifies to: 
$$\mathbb{E}_{S_n}[q(0|x)]c(f_0,g)+(1-\mathbb{E}_{S_n}[q(0|x)])c(f_1,g)\leq B.$$ 
Rearranging terms,
$$
\mathbb{E}_{S_n}[q(0|x)] (c(f_0,g)-c(f_1,g)) \leq B - c(f_1,g).
$$
The budget constraint thus depends on 3 variables: $\mathbb{E}_{S_n}[q(0|x)]$, $c(f_1,g)$ and $c(f_0,g)$. 
To meet the constraint, we would like to have (a) low-cost model and gating (small $c(f_1,g)$ and $c(f_0,g)$); and (b) small fraction of examples being sent to the high-accuracy model (small $\mathbb{E}_{S_n}[q(0|x)]$).
We can therefore split the budget constraint into two separate objectives: (a) ensure low-cost through penalty $\Omega(f_1,g)$, (b)  Ensure only $\text{P}_{\text{full}}$ fraction of examples reach $f_0$. 
\begin{align*} \tag{OPT2}\label{eq:OPT2}
\hspace{-.15cm}
\min_{f_1 \in {\cal F}, g \in {\cal G}, q}
&\mathbb{E}_{S_n}\sum\limits_{z}[ q(z|x)\ell(f_z(x),y)] + D(q(\cdot|x),g(x)) +\Omega(f_1,g)\\
\text{subject to: } & \mathbb{E}_{S_n}[q(0|x)] \leq P_{\text{full}} 
\end{align*}
Compared to \eqref{eq:OPT1}, \eqref{eq:OPT2} is much more amenable to alternating minimization.

\noindent
{\bf General Framework:}
We presented the case for a single stage approximation system. However, it is straightforward to recursively continue this process. We can then view the composite system $f_0 \triangleq (g,f_1, f_0)$ as a black-box predictor and train a new pair of gating and prediction models to approximate the composite system. Depending on the parameterization of models and choice of the distance function $D(q(\cdot|x), g(x))$, our general framework encompasses a large number of models.

\subsection{Two Problems}
We consider two classes of problems involving prediction under a budget. The main focus of this paper is on the first problem.

\subsubsection{Adaptive feature acquisition problems}  
In these applications each feature, say $\alpha$, is associated with an acquisition cost, say  $c_{\alpha}$ during test-time. Often we can train a complex, high-accuracy model $f_0$ that requires the use of all the features for test examples. So the cost for using $f_0$ is  $C=\sum_{\alpha} c_\alpha$, which can be prohibitively high.
The task of adaptive feature acquisition is to acquire features in an input-dependent manner so as to reduce average feature costs while maintaining high prediction accuracy. 
Intuitively, ``easy'' examples can be correctly classified using only a few cheap features as they are sent to $f_1$ while ``hard'' examples require more expensive features and are sent to $f_0$. The gating is performed by $g$, which should itself be low-cost.

Given an example $x$, we assume that once it pays the cost to acquire a feature, its value can be stored in memory; its subsequent use does not incur additional cost. Thus the cost of predicting $x$ using the $(g,f_1,f_0)$ system is the sum of the acquisition cost of \emph{unique} features required by the system. 
This means that the cost for feature re-use is zero, and it is advantageous if gating and prediction models share common features. So the fundamental question we need to answer is:

{\emph{How to choose a common, sparse (low-cost) subset of features on which both $g$ and $f_1$ operate, such that $g$ can effective gate examples between $f_1$ and $f_0$ for accurate prediction?}}

This is a hard combinatorial problem. The main contribution of our paper is to address it using the general optimization framework of \eqref{eq:OPT1} and \eqref{eq:OPT2}.

To see the connection, let indicator variables $V_{\alpha}(x),W_{\alpha}(x)$ denote whether or not the feature $\alpha$ is utilized for example $x$ by $f_1$ and $g$, respectively. The LHS of the constraint in \eqref{eq:OPT1} is:
\begin{align}\label{eq:cost0}
\mathbb{E}_{S_n}\mathbb{E}_{q(|x)}[c(f_z,g,x)] = \frac{C}{n}\sum_{i=1}^n q(0| x_i) 
+\frac{1}{n}\sum_{i=1}^n (\sum_{\alpha} c_\alpha V_{\alpha}(x_i)\vee W_{\alpha}(x_i))q(1| x_i).
\end{align}
In special cases where the feature utilization is independent of the instance, i.e. $V_{\alpha}(x)=V_{\alpha}, W_{\alpha}(x)=W_{\alpha}, \forall x$, we can model the cost to be
\begin{align}\label{eq:cost}
\mathbb{E}_{S_n}\mathbb{E}_{q(|x)}[c(f_z,g,x)] = C \frac{1}{n}\sum_{i=1}^n q(0 | x_i) +
(\sum_{\alpha} c_\alpha V_{\alpha}\vee W_{\alpha})\left(1- \frac{1}{n}\sum_{i=1}^n q(0| x_i)\right) 
\end{align}
Following the reduction of \eqref{eq:OPT1} to \eqref{eq:OPT2}, we can decouple the feature cost penalty from the proportion of examples reaching the full predictor. In particular this leads to a penalty for cost of feature utilization through: $\Omega(f_1,g)=\sum_{\alpha}c_\alpha \|V_{\alpha}+W_{\alpha}\|_0$, which can be further relaxed using a group-sparse norm to obtain a convex surrogate as we will see shortly in \eqref{eq:OPT4}.  

\subsection{Local Constrained Systems}
In this scenario we have a remote, high-accuracy model $f_0$. $g$ and $f_1$ are deployed locally, which are limited by memory, battery and CPU power. Sending examples to $f_0$ incurs communication and latency cost. So it is desirable to predict as many examples as possible locally using $f_1$. Only the ``hard'' examples should be sent to $f_0$. 
This situation is aligned with \eqref{eq:OPT2} and the cost function is similar to Eq~\ref{eq:cost}. Specifically, local costs can be accounted through model complexity and penalizing any computations exclusive to either gating or prediction models. For remote computation cost we constrain the average number of examples reaching $f_0$ as in \eqref{eq:OPT2}.

We will next list different options for the loss functions, distance measures and parameterized model classes that we experiment with in this paper. 

\subsection{Gating Approximations \& Losses}
We exclusively employ logistic loss functions, although our framework allows for other loss models. Specifically, as is well known this entails a probability model for posterior distribution for the outputs. We restrict our attention here to binary losses. In this context prediction and gating probabilities are of the form:
\vspace{-0.4cm}
$$
\text{Pr}(y|f_z)=\frac{1}{1+e^{-yf_z(x)}},\,\,
\text{Pr}(z=0|;g)=\frac{1}{1+e^{-g(x)}}.
$$
\paragraph{Loss Function:} Our loss function is $\ell(f_z(x),y)=\log(1+\exp(-yf_z(x))$. We briefly motivate our choice of this loss function. First the excess loss, ${\partial \ell}_{0,1} = \ell(f_1(x),y)-\ell(f_0(x),y)$ is the deviance between low-cost model and the full ``saturated'' model:
$$
\text{Dv}_{f_0,f_1}\triangleq -\log \frac{P(y|f_1)}{P(y|f_0)} = \log \frac{1+e^{-yf_0(x)}}{1+e^{-yf_1(x)}}\triangleq{\partial \ell}_{0,1}
$$
The deviance as a measure of goodness of fit has been widely studied~\cite{app_logistic}. At a fundamental level it is a likelihood ratio test (LRT) for identifying regions of input space with good fit. We describe a few benefits.

First, we only penalize the low-cost predictor relative to the performance of the high-accuracy model. Thus the low-cost predictor is unconstrained in the poorly performing regions of $f_0$. Furthermore, margin of $f_0$ behaves as a slack-variable that $f_1$ attempts to not exceed. 
Second, note that the posterior probability in regions of high-confidence are likely to behave smoothly and amenable to approximation with low-cost models thus allowing for natural partitions of input space. Third, from a LRT viewpoint our choice implicitly functions as a hypothesis test for the gating function for resolving between $f_0$ and $f_1$. Viewed from this perspective, since thresholding the likelihood ratio is Bayes optimal under a variety of cost functions, our choice implicitly encodes optimal target distributions for gating to mimic. 

\paragraph{Gating Approximation:}
A natural choice for an approximation metric is the Kullback-Leibler (KL) divergence although other choices are possible. The KL divergence between two distributions, $r,s$, is given by $D_{KL}(r\|s) = \sum_z r(z) \log(r(z)/s(z))$. We can better insight for this choice we view $z \in \{0,1\}$ as a latent variable and consider the composite system $\Pr(y|x) = \sum_z \Pr(z|x;g) \Pr(y|x,f_z)$. A standard application of Jensen's inequality reveals that, 
\begin{align*}
-\log(\Pr(y|x)) \leq \mathbb{E}_{q(z|x)} \ell(f_z(x),y) + D_{KL}(q(z|x)\|\Pr(z|x;g))
\end{align*}
This implies that our objective attempts to bound the loss of the composite system and the constraints serve to enforce budget limits on the composite system. 

\paragraph{Other Symmetrized metrics:} KL divergence is not symmetric and leads to widely different properties in terms of approximation. We also consider a symmetrized metric:
\vspace{-0.2cm}
$$
D(r(z),s(z)) = \left (\log \frac{r(0)}{r(1)} - \log \frac{s(0)}{s(1)}\right )^2
$$
This metric can be viewed intuitively as a regression of $g(x)=\log(\Pr(1|g;x)/\Pr(0|g;x)$ against the observed log odds ratio of $q(z|x)$.

The main advantage of using KL is that optimizing w.r.t. $q$ can be solved in closed form \eqref{eq:OPT3}. The disadvantage we observe is that in some cases, the loss for minimizing w.r.t. $g$, which is a weighted sum of log-losses of opposing directions, becomes quite flat and difficult to optimize especially for linear gating functions. The symmetrized measure, on the other hand, makes the optimization w.r.t. $g$ better conditioned as the gating function $g$ fits directly to the log odds ratio of $q$. However, the disadvantage of using the symmetrized measure is that optimizing w.r.t. $q$ no longer has closed form solution; furthermore, it is even non-convex. We offer an ADMM approach for $q$ optimization (ref. Appendix).


\paragraph{Relationship to Hierarchical Mixture of Experts (HME):} Our composite system is related to HME~\cite{Jordan:1994:HME:188104.188106}. HME casts the problem as learning the composite system based on max-likelihood estimation of models. While HME is closely related there are conceptual differences. First, a major difference is that we have budget constraints that HME does not address. A fundamental aspect is the significant asymmetry in our setup. Our full model is assumed to be highly accurate (albeit with high-cost) unlike the HME scenario, where the component parts satisfy similar roles. This asymmetry leads us to propose a bottom-up strategy where the high-accuracy predictor can be separately estimated. This asymmetry ultimately results in posing our objective as a more direct empirical loss minimization problem.

\paragraph{Joint Convexity:}
We point out in passing that our objective function of \eqref{eq:OPT1} is jointly convex in the space of probability densities. 
\begin{lemma}
	The objective function
	${\cal L}=\sum_z q(z|x) \ell(f_z(x),y) + D_{KL}(q(z|x) \| \Pr(z|x;g))$
	is jointly convex in $q(z|x),\,\Pr(y | x;f_1),\,\Pr(z | x;g)$ if $\ell(f_1(x),y) = -\log (\Pr(y | x;f_1))$.  
\end{lemma}
The proof of the result directly follows from \cite{cover}. As a consequence \eqref{eq:OPT1},\eqref{eq:OPT2} are convex problems if the sets $\{\Pr(y | x;f_1),\,f_1 \in {\cal F}\}$, $\{\Pr(z | x;g),\,g \in {\cal G}\}$ are convex subsets and the cost constraint in \eqref{eq:OPT1} is a convex or the penalty function in \eqref{eq:OPT2} is convex. In the sequel we consider both parametric and non-parametric classes of predictors. While convexity does not hold in general for parametric classes, it holds for non-parametric universal function classes. 

\subsection{Parameterized Model Classes}

To be concrete, we instantiate our general framework \eqref{eq:OPT1} into two algorithms via different parameterizations of $g,f_1$.

\paragraph{Linear class:} Let $g(x) = g^Tx$ and $f_1(x)=f_1^Tx$ be linear classifiers. A feature is used if the corresponding component is non-zero: $V_\alpha=1$ if $f_{1,\alpha} \neq 0$, and $W_\alpha=1$ if $g_\alpha \neq 0$. 

\paragraph{Non-parametric class:}
We can also consider non-parametric family of classifiers such as gradient boosted trees \cite{Friedman00greedyfunction}: $g(x) = \sum_{t=1}^{T} g^t(x)$ and $f_1(x)=\sum_{t=1}^{T} f_1^t(x)$, where $g^t$ and $f_1^t$ are limited-depth regression trees. 
Since the trees are limited to low depth, we assume that the feature utility of each tree is example-independent: $V_{\alpha,t}(x)\approxeq V_{\alpha,t}, W_{\alpha,t}(x)\approxeq W_{\alpha,t},\forall x$. $V_{\alpha,t}=1$ if feature $\alpha$ appears in $f_1^t$, otherwise $V_{\alpha,t}=0$, similarly for $W_{\alpha,t}$.
\begin{algorithm}[h]
   \caption{\textsc{DynaMod-Lin}}
   \label{alg:dynamod-lin}
\begin{algorithmic}
   \STATE {\bfseries Input:} $(x^{(i)},y^{(i)}),\text{P}_{\text{full}}, \gamma$
   \STATE Train $f_0$. Initialize $g, f_1$.
   \REPEAT
   \STATE Solve (OPT3) for $q$ given $g, f_1$. 
   \STATE Solve (OPT4) for $g, f_1$ given $q$.
   \UNTIL{convergence}
\end{algorithmic}
\end{algorithm}
\section{Algorithms}
Depending on the distance function and parameterization, our general framework leads to different algorithms. We propose two: \textsc{DynaMod-lin} for the linear class and \textsc{DynaMod-Gbrt} for the non-parametric class. Both of them solve the adaptive feature acquisition problem. Both of them use the KL-divergence as distance measure. We also provide a third algorithm \textsc{DynaMod-Lstsq} that uses the symmetrized squared distance in the Appendix. All of the algorithms perform alternating minimization of \eqref{eq:OPT2} over $q,g,f_1$. 
Note that convergence of alternating minimization follows as in \cite{NIPS2007_3170}. Common to all of our algorithms, we use two parameters to control cost: $\text{P}_{\text{full}}$ and $\gamma$. In practice they are swept to generate various cost-accuracy tradeoffs and we choose the best one satisfying the budget $B$ using validation data.

\subsection{\textsc{DynaMod-lin}} The minimization for $q$ solves the following problem:
\begin{equation}
\begin{array}{rlll}\tag{OPT3}\label{eq:OPT3}
\displaystyle \min_{q} &  \multicolumn{2}{l}{\frac{1}{N} \sum_{i=1}^{N} \left [(1-q_i)A_i+q_iB_i - H(q_i)\right ]} \\
\textrm{s.t.} &  \frac{1}{N} \sum_{i=1}^{N} q_i \leq \text{P}_{\text{full}}, \end{array}
\end{equation}
where we have used shorthand notations $q_i = q(z=0|x^{(i)})$, $H(q_i)=-q_i\log(q_i)-(1-q_i)\log(1-q_i)$, $A_i=\log(1+e^{-y^{(i)}f_1^Tx^{(i)}})+\log(1+e^{g^Tx^{(i)}})$ and $B_i=-\log p(y^{(i)}|z^{(i)}=0;f_0)+\log(1+e^{-g^Tx^{(i)}})$.
This optimization has a closed form solution: $q_i = 1/(1+e^{B_i-A_i+\beta})$ for some constant $\beta$ such that the constraint is satisfied. This optimization is also known as I-Projection in information geometry \cite{NIPS2007_3170}.
Having optimized $q$, we hold it constant and minimize with respect to $g,f_1$ by solving the problem \eqref{eq:OPT4}, where we have relaxed the non-convex cost $\sum_{\alpha}c_\alpha \|V_{\alpha}+W_{\alpha}\|_0$ into a $L_{2,1}$ norm for group sparsity and a tradeoff parameter $\gamma$ to make sure the feature budget is satisfied. Once we solve for $g,f_1$, we can hold them constant and minimize with respect to $q$ again. \textsc{DynaMod-Lin} is summarized in Algorithm~\ref{alg:dynamod-lin}. 
\begin{equation}\tag{OPT4}\label{eq:OPT4}
\displaystyle \min_{g,f_1} \frac{1}{N} \sum_{i=1}^{N} \left [(1-q_i) \left (\log(1+e^{-y^{(i)}f_1^Tx^{(i)}})+\log(1+e^{g^Tx^{(i)}}) \right )+q_i \log(1+e^{-g^Tx^{(i)}})\right ] + \gamma \sum_{\alpha} \sqrt{g_{\alpha}^2+f_{1,\alpha}^2}.
\end{equation}

\subsection{\textsc{DynaMod-Gbrt}}
The optimization over $q$ still solves \eqref{eq:OPT3}. We modify $A_i=\log(1+e^{-y^{(i)}f_1(x^{(i)})})+\log(1+e^{g(x^{(i)})})$ and $B_i=-\log p(y^{(i)}|z^{(i)}=0;f_0)+\log(1+e^{-g(x^{(i)})})$. The main difference with the linear setting is the optimization over $g,f_1$. We adopt a greedy approach to function approximation \cite{Friedman00greedyfunction}. Denote loss: 
\begin{align*}
\ell(f_1,g) =\frac{1}{N} \sum_{i=1}^{N} \Bigg[q_i \log(1+e^{-g(x^{(i)})}) + (1-q_i)\cdot 
\left (\log(1+e^{-y^{(i)}f_1(x^{(i)})})+\log(1+e^{g(x^{(i)})}) \right )\Bigg],
\end{align*}
which is essentially the same as the first part of the objective in (OPT4). 

Following the approximations done in \textsc{GreedyMiser} \cite{DBLP:conf/icml/XuWC12}, we can minimize the loss $\ell(f_1,g)$ and feature cost by employing an impurity function as follows $I(f_1^t)$:
\begin{equation} \label{eq:impurity_h}
\frac{1}{2}\sum_{i=1}^{N}(-\frac{\partial \ell(f_1,g)}{\partial f_1(x^{(i)})} - f_1^t(x^{(i)}))^2+\gamma\sum_{\alpha} u_\alpha c_\alpha V_{\alpha,t},
\end{equation}
where $u_\alpha \in \{0,1\}$ indicates if feature $\alpha$ has already been used in previous trees ($u_\alpha=0$), or not ($u_\alpha=1$). Similarly $I(g^t)$:, 
\begin{equation} \label{eq:impurity_g}
\frac{1}{2}\sum_{i=1}^{N}(-\frac{\partial \ell(f_1,g)}{\partial g(x^{(i)})} - g^t(x^{(i)}))^2+\gamma\sum_{\alpha} u_\alpha c_\alpha W_{\alpha,t}.
\end{equation}
This impurity balances the need to minimize loss and re-using the already acquired features. Classification and Regression Tree (CART) \cite{breiman1984classification} can be used to construct decision trees with such an impurity function. 

Interestingly, if $\text{P}_{\text{full}}$ is set to 0, all the examples are forced to $f_1$, then \textsc{DynaMod-Gbrt} exactly recovers the \textsc{GreedyMiser}. In this sense, \textsc{GreedyMiser} is a special case of our algorithm. 

\begin{algorithm}[tb]
   \caption{\textsc{DynaMod-Gbrt}}
   \label{alg:dynamod-gbrt}
\begin{algorithmic}
   \STATE {\bfseries Input:} $(x^{(i)},y^{(i)}),\text{P}_{\text{full}}, \gamma$
   \STATE Train $f_0$. Initialize $g,f_1$.
   \REPEAT
   \STATE Solve (OPT3) for $q$ given $g, f_1$. 
   \FOR{$t=1$ {\bfseries to} $T$}
   		\STATE Find $f_1^t$ using CART to minimize \eqref{eq:impurity_h}.
   		\STATE $f_1=f_1 + f_1^t$.
   		\STATE For each feature $\alpha$ used, set $u_\alpha=0$.
   		\STATE Find $g^t$ using CART to minimize \eqref{eq:impurity_g}.
   		\STATE $g=g+ g^t$.
   		\STATE For each feature $\alpha$ used, set $u_\alpha=0$.
   \ENDFOR
   \UNTIL{convergence}
\end{algorithmic}
\end{algorithm}

\section{Experiments}
We will highlight several important aspects of our method through experiments on real and synthetic datasets. {\bf (a)} Our experiments will show that our method outperforms other bottom-up procedures such as greedy and RF pruning on both real and synthetic datasets. {\bf (b)} It significantly outperforms state-of-the-art bottom-up and top-down methods on several datasets. {\bf (c)} We will show that it is capable of adaptively approximating powerful classifiers including RBF-SVM, Random Forest(RF), XGBoost\cite{Chen:2016:XST:2939672.2939785}. {\bf (d)} It can be utilized for a variety of tasks such as ranking and classification. {\bf (e)} It can account for different types of costs including uniform and non-uniform feature costs as well as model complexity costs. 



We experiment with a synthetic dataset and four other datasets. Experimental details can be found in Appendix. The first three datasets have unknown feature acquisition costs so we assign unit cost to all features. When features have unit cost, the adaptive feature acquisition problem is an adaptive sparse feature selection problem. 
%
The fourth dataset has real acquisition costs measured in terms of CPU time. 

\begin{figure}[htbp]
\centering
\includegraphics[width=0.45\textwidth]{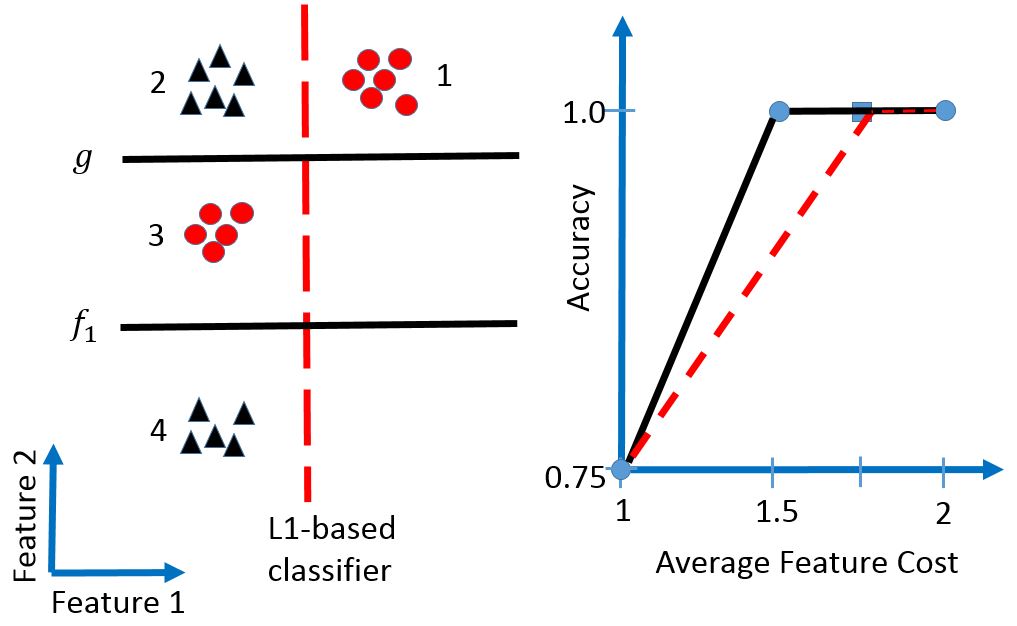}
\caption{A 2-D synthetic example for adaptive feature acquisition. On the left: data distributed in four clusters. The two features correspond to x and y coordinates, respectively. The examples can be correctly classified if both features are acquired. On the right: accuracy-cost tradeoff curves. Our algorithm can recover the optimal adaptive system whereas a L1-based approach cannot.} \label{fig:synthetic}
\end{figure}

\paragraph{Synthetic dataset:} We consider a 2D dataset as shown in Figure \ref*{fig:synthetic}(left). The examples are distributed in four clusters where cluster 1 and 3 are of class 0 and cluster 2 and 4 are of class 1. Cluster 2 and 1 cannot be distinguished by acquiring feature 2 alone and cluster 2,3 and 4 cannot be distinguished by acquiring feature 1 alone. If both feature 1 and 2 have unit cost, a complex classifier that acquires both features can achieve full accuracy at the cost of 2. 

Consider an adaptive system where the gating function is the  solid line separating clusters 1 and 2 from 3 and 4, sending the first two clusters to the full classifier and the last two clusters to a partial classifier. The low-cost prediction model $f_1$ is the solid line that correctly classifies clusters 3 and 4. So all of the examples are correctly classified by the adaptive system; yet only feature 2 needs to be acquired for cluster 3 and 4 so the overall average feature cost is 1.5, as shown by the solid curve in the accuracy-cost tradeoff plot on the right of Figure \ref*{fig:synthetic}. The greedy approach first performs a L1-regularized logistic regression which would give the vertical red dashed line, separating cluster 1 from the others. So feature 1 is acquired for both $g$ and $f_1$. The best such an adaptive system can do is to send cluster 1 to the $f_1$ and the other three clusters to the complex classifier $f_0$, incurring an average cost of 1.75. Thus the greedy approach leads to a sub-optimal accuracy-cost tradeoff as shown by the dashed curve on the right of Figure \ref*{fig:synthetic}. Our algorithm \textsc{DynaMod-Lin}, on the other hand, optimizing between $q,g,f_1$ in an alternating manner, is able to recover the optimal adaptive system. Please refer to the Appendix for more detailed information. 

\paragraph{Letters Dataset \cite{UCI_repository}:} This letters recognition dataset contains 20000 examples with 16 features, each of which is assigned unit cost. We use two different pre-trained classifiers as $f_0$: a RBF SVM and a random forest (RF) of 500 trees. Given each $f_0$, we compare the greedy approach (See Appendix), \textsc{DynaMod-Lin} and \textsc{DynaMod-Gbrt}. 
This dataset serves to illustrate: A) our algorithms can work with general classifiers as input and achieve high performance; the red curves correspond to using RBF-SVM as $f_0$ and the black ones corresponds to RF. \textsc{DynaMod-Gbrt} in particular can well maintain high accuracy for both input classifiers while reducing cost. Thus in practice we can apply our algorithm to the best $f_0$ to achieve the highest performance (RBF-SVM in this case). B) using non-parametric class leads to significant better performance than the linear one; \textsc{DynaMod-Gbrt} clearly outperforms \textsc{DynaMod-Lin} in both cases. This shows the advantage of nonlinear classifiers on real dataset. C) in the linear setting our algorithm outperforms the greedy approach (denoted as L1 in the figure). The greedy approach first selects a subset of features via L1 regularized logistic regression; then learns $f_1$ based on the feature subset; then learns $g$ based on the output of $f_1$ and $f_0$. See more details in Appendix. Again, this confirms the intuition we have in the synthetic experiment on real data. The main reason that \textsc{DynaMod-Lin} performs better is that it can iteratively select the common subset of features jointly for $g$ and $f_1$. D) With a comparable model size, \textsc{GreedyMiser} cannot achieve the high accuracy as \textsc{DynaMod-Gbrt} and the gap is quite significant. This precisely shows the advantage of our bottom-up approach. \textsc{DynaMod-Gbrt} is able to leverage the high-accuracy model of RBF-SVM or RF whereas a top-down, ``tree-growing'' approach such as \textsc{GreedyMiser} cannot. 
\begin{figure}[htbp]
\centering
\includegraphics[width=0.45\textwidth]{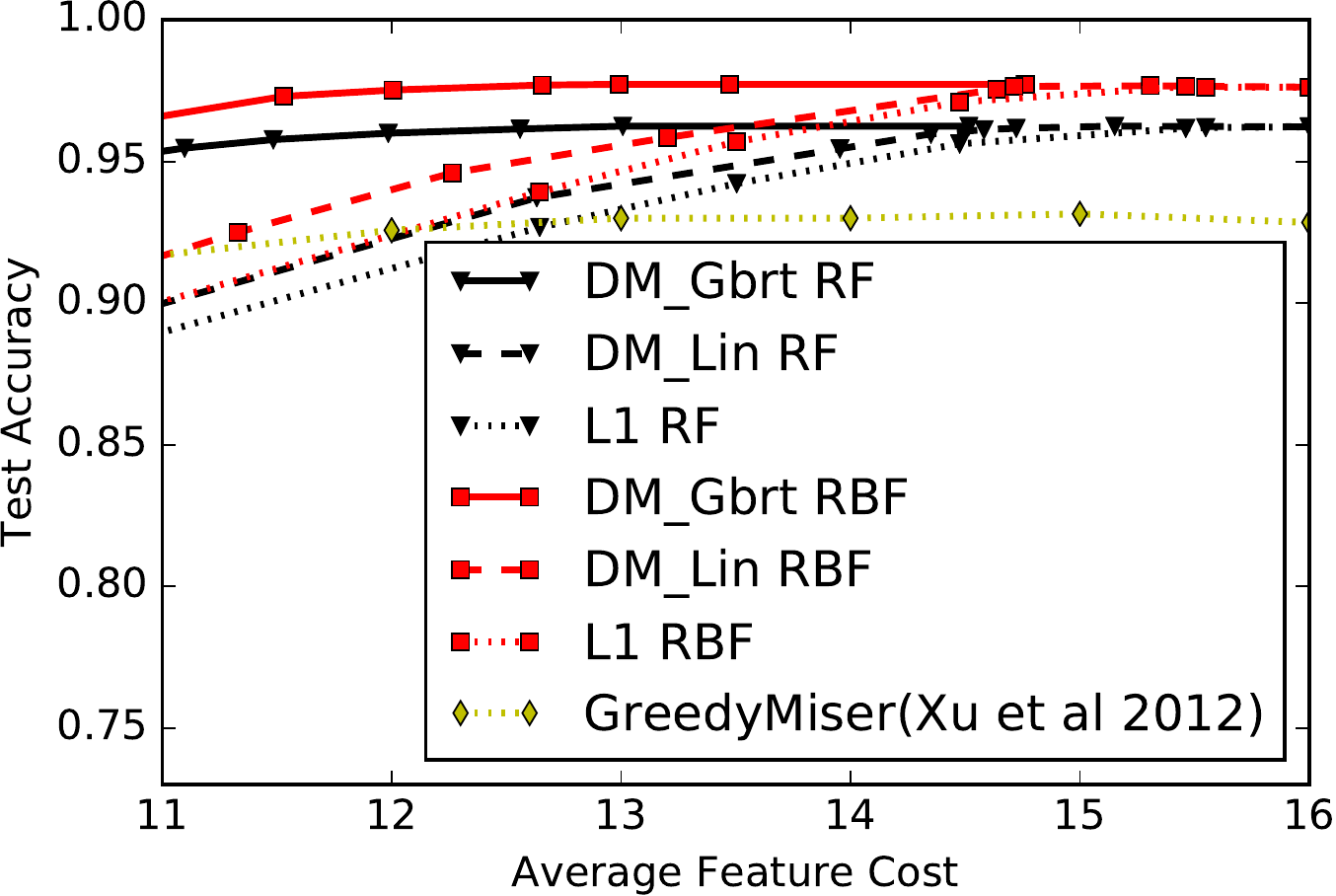}
\caption{Compare the greedy L1 approach, \textsc{DynaMod-Lin} and \textsc{DynaMod-Gbrt} based on RBF-SVM and RF as input $f_0$'s. Our Method can approximate powerful classifiers and is superior to other Bottom-up methods as well as \textsc{GreedyMiser}} \label{fig:letters}
\end{figure}

\begin{figure*}[htb!]
\vspace{-0.05in}
\centering
\subfigure[MiniBooNE]{\includegraphics[width=.32\linewidth,height=.29\linewidth]{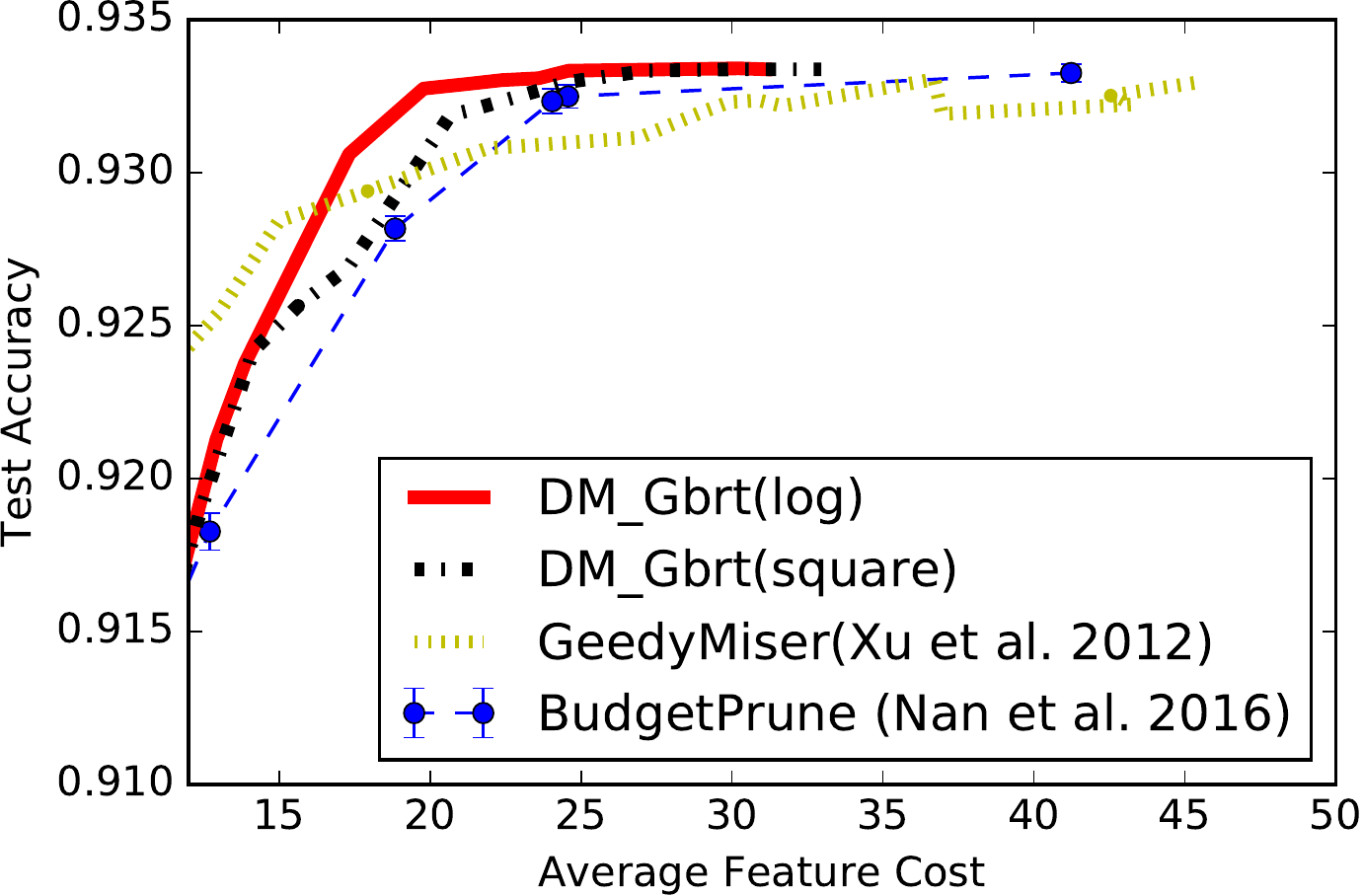}}
\vspace{-0.1in}
\subfigure[Forest Covertype]{\includegraphics[width=.32\linewidth,height=.29\linewidth]{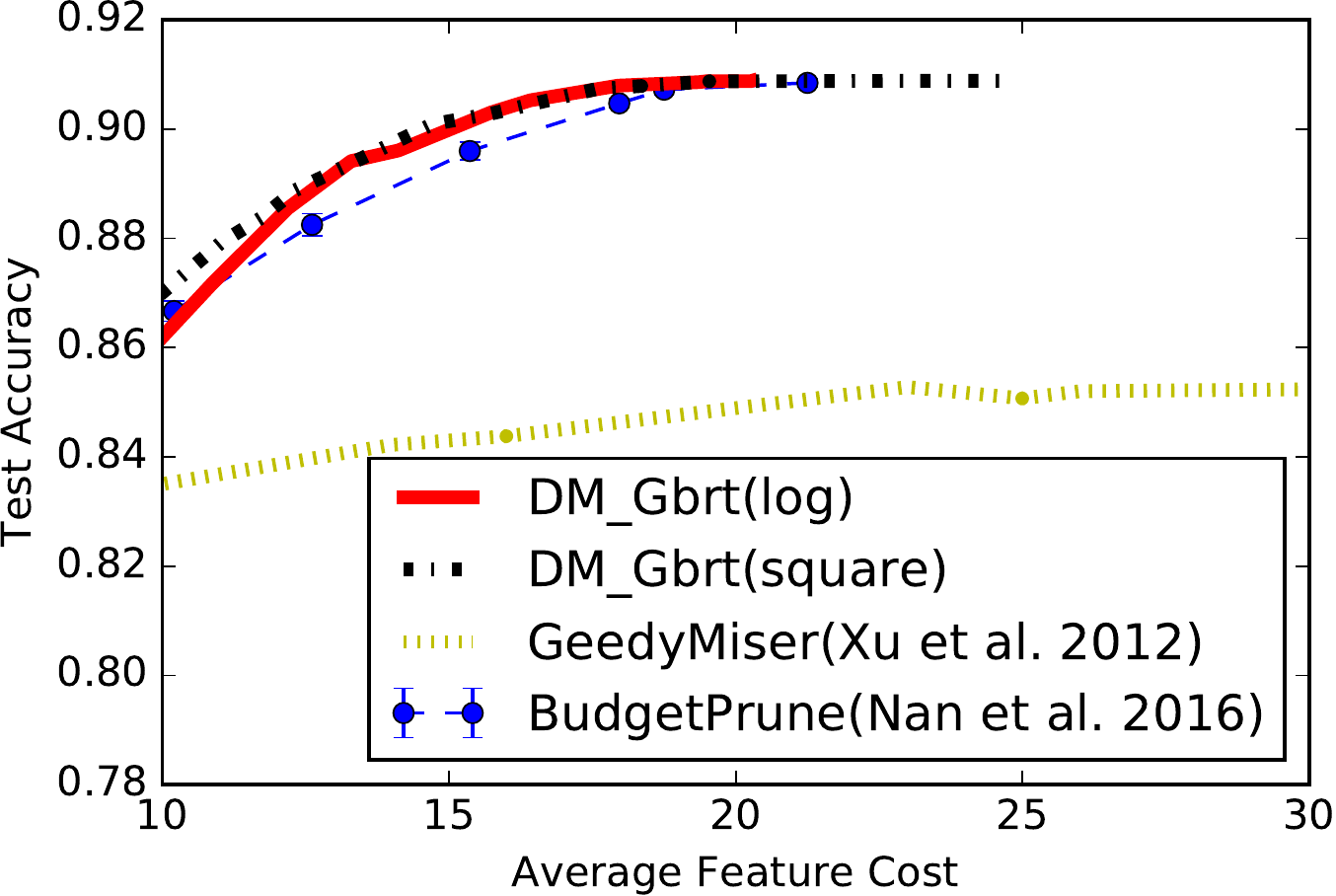}}
\subfigure[Yahoo! Rank]{\includegraphics[width=.32\linewidth,height=.285\linewidth]{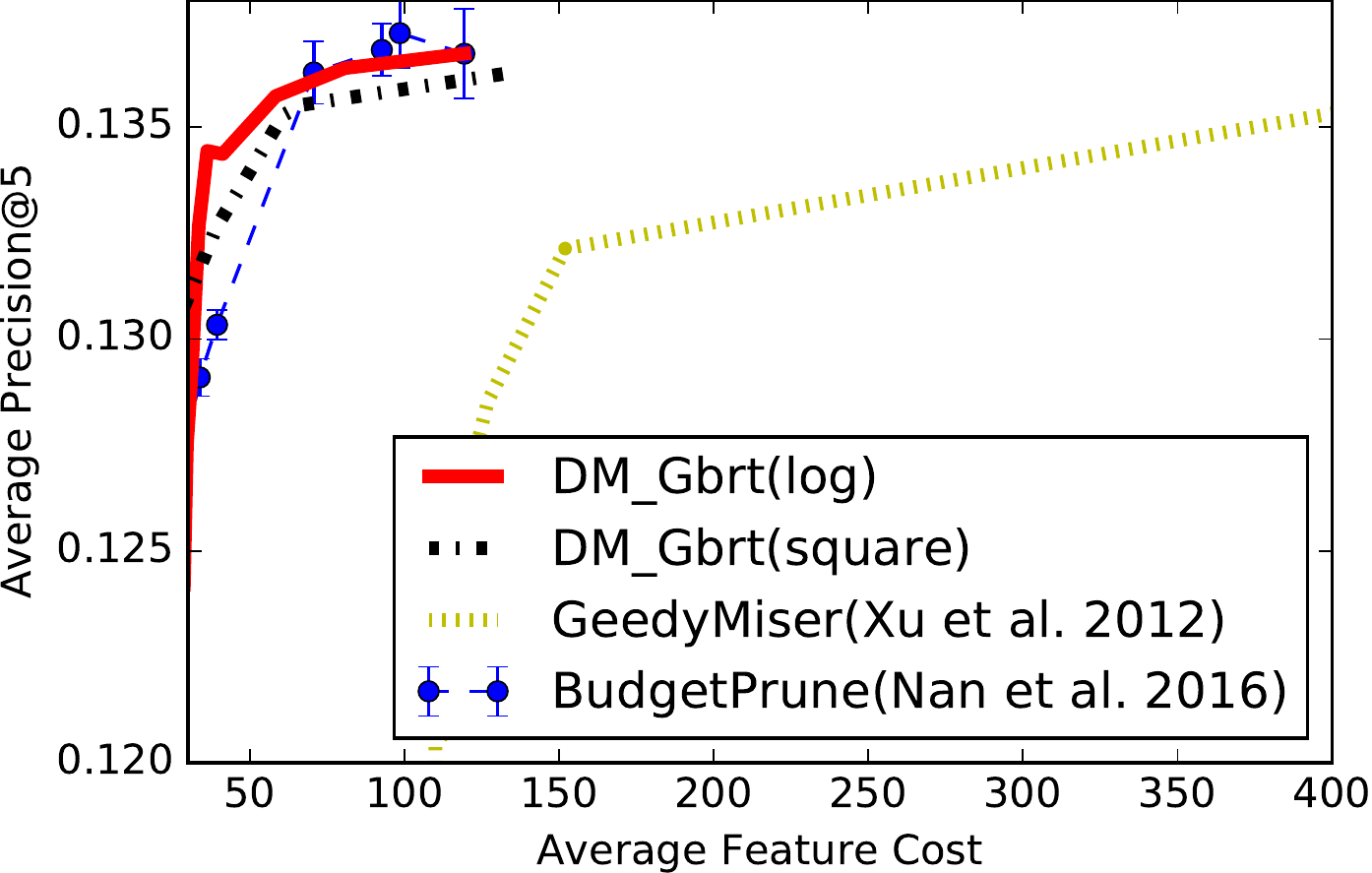}}
\caption{Comparison of \textsc{DynaMod-Gbrt} against \textsc{GreedyMiser} \cite{DBLP:conf/icml/XuWC12} and \textsc{BudgetPrune} \cite{NanNIPS2016}. \textsc{DynaMod-Gbrt} achieves high accuracy using less feature cost than the  state-of-the-art Bottom-up (BudgetPrune) and Top-down (\textsc{GreedyMiser}) methods. Experiments on Scene 15 and CIFAR for feature costs were also conducted. Both resulted in marginal improvements for different reasons. Scene 15 is a relatively small dataset. For CIFAR, GBRT has state-of-art performance~\cite{xu2013cost} and greedy miser has excellent cost performance. \textsc{GreedyMiser} is a special case of our method.}
\label{fig:experiments}
\end{figure*}

\paragraph{MiniBooNE Particle Identification and Forest Covertype Datasets \cite{UCI_repository}:} Feature costs are uniform in both datasets. We apply \textsc{DynaMod-Gbrt} using two distance functions: KL-divergence (denoted as log in the figures) and the symmetrized metrics (denoted as square in the figures) as described in Section 3.0.1 for $g$ and $f_1$. We choose the unpruned RF of \textsc{BudgetPrune} \cite{NanNIPS2016} as the complex classifier $f_0$ as it has the highest accuracy. As shown in (a) and (b) of Figure \ref{fig:experiments}, \textsc{DynaMod-Gbrt} clearly achieves better accuracy-cost tradeoff than \textsc{BudgetPrune} and \textsc{GreedyMiser} \cite{DBLP:conf/icml/XuWC12}. Furthermore, the KL-divengence distance seems to perform better than the symmetrized distance. 

\paragraph{Yahoo! Learning to Rank\cite{YahooChallenge2010}:} This ranking dataset consists of 473134 web documents and 19944 queries. Each example is associated with features of a query-document pair together with the relevance rank of the document to the query. There are 519 such features in total; each is associated with an acquisition cost in the set \{1,5,20,50,100,150,200\}, which represents the units of CPU time required to extract the feature and is provided by a Yahoo! employee. The labels are binarized into relevant or not relevant. The task is to learn a model that takes a new query and its associated documents and produce a relevance ranking so that the relevant documents come on top, and to do this using as little feature cost as possible. The performance metric is Average Precision @ 5 following \cite{NanNIPS2016}. Again we take the unpruned RF of \textsc{BudgetPrune} as input $f_0$. As shown in (c) of Figure \ref{fig:experiments}, \textsc{BudgetPrune} has a higher accuracy around cost 100 because the initial pruning improves the generalization ability. But in the cost region of 40-80, the \textsc{DynaMod-Gbrt} KL-divergence maintains much better accuracy than \textsc{BudgetPrune}. 

\paragraph{Local Constrained Systems:} Finally, we consider the scenario of a local-remote system. We assume the remote system has a complex classifier $f_0$, in our experiments 500 gradient boosted trees using XGBoost \cite{Chen:2016:XST:2939672.2939785}. We assume the local system has limited memory/power that it can only deploy a small number of trees for $g$ and $f_1$, in our experiments 10 XGBoost trees. The task is to learn $g$ and $f_1$ given $f_0$ such that the local classifier $f_1$ can process as many examples as possible without having to communicate to the cloud $f_0$ yet still maintains high accuracy. We start with training 10 XGBoost trees as $f_1$. A confidence-based approach selects a threshold $\tau$ and sends all examples with high margin $|f_1(x)|>\tau$ to $f_1$ and otherwise to $f_0$. So $f_1$ serves as the gating function $g$ to meet budget constraint. We can also introduce a separate $g$ that shares the same tree structure as $f_1$ but with different leaf weights. This provides additional flexibility yet does not incur much overhead. We can then use the greedy approach,\textsc{DynaMod-Lin} and linear \textsc{DynaMod-Lstsq} (See Appendix) to learn the leaf weights $g$ and $f_1$.  Table\ref{table:local_remote} shows a comparison of different approaches. The second and the third columns show the test accuracies of the XGBoost with 500 and 10 trees, respectively. The 4th column is the desired accuracy level. Columns 5-8 show the percentages of test examples that have to be sent to the remote system ($f_0$) in order to achieve the desired accuracy as in column 4. The smaller the percentage, the better the adaptive system performs. The uniform approach gates examples randomly between $f_0$ and $f_1$.
We see that confidence-based method is often quite effective as it sends less than 50\% of the examples to the remote system while losing only one percent in accuracy. Greedy often does worse than confidence-based method. We believe it is due to the fact that the margin output of $f_1$ is so effective as a gating statistics that re-learning the weights for $g$ actually gets worse results. Nevertheless, we see that our \textsc{DynaMod-Lstsq} can improve upon the confidence-based adaptive system by adjusting the weights $g$ and $f_1$.

\begin{table}[h]
\centering
\caption{Percentages sent to the remote for a given accuracy in adaptive local-remote systems. Lower is better. (best highlighted in red)}
\label{table:local_remote}
\resizebox{\columnwidth}{.15\linewidth}{
\begin{tabular}{|l|l|l|l|l|l|l|l|}
\hline
                         & $f_0$ accu                & $f_1$ accu                & target accu & uniform & confidence & greedy & DM-Lstsq \\ \hline
\multirow{2}{*}{Mbne}    & \multirow{2}{*}{0.942} & \multirow{2}{*}{0.921} & 0.94       & 90       & 18               & 18     & \textcolor{red}{15}            \\ \cline{4-8} 
                         &                        &                        & 0.938      & 81       & 13               & 12     & \textcolor{red}{11}            \\ \hline
\multirow{2}{*}{Foresst} & \multirow{2}{*}{0.885} & \multirow{2}{*}{0.818} & 0.88       & 93       & 43               & \textcolor{red}{42}     & 43            \\ \cline{4-8} 
                         &                        &                        & 0.87       & 78       & 34               & 33     & \textcolor{red}{32}            \\ \hline
\multirow{2}{*}{Cifar}   & \multirow{2}{*}{0.749} & \multirow{2}{*}{0.675} & 0.74       & 88       & 65               & 67     & \textcolor{red}{63}            \\ \cline{4-8} 
                         &                        &                        & 0.73       & 74       & 49               & 52     & \textcolor{red}{43}            \\ \hline
\end{tabular}
}
\end{table}
%
\section{Conclusions}
We presented a dynamic model selection approach to account for a diverse range of costs including feature costs, sensor acquisition costs and local computational costs that arise in various applications. 
At test-time our method uses a gating function to identify a prediction model among a collection of models that is adapted to the input. The overall goal is to reduce costs without sacrificing accuracy. We learn gating and prediction models on fully labeled training data by means of a bottom-up strategy that recursively trains low-cost models to approximate high-complexity models in regions where low-cost models suffice. On a number of benchmark datasets our method outperforms state-of-the-art achieving superior accuracy for the same cost budget. Our conclusion is that a bottom-up strategy leads to superior performance in regimes where we wish to reduce costs but still maintain high-accuracy.

\subsubsection*{Acknowledgments}

Feng Nan would like to thank Dr Ofer Dekel for ideas and discussions on resource constrained machine learning during an internship in Microsoft Research in summer 2016. Familiarity and intuition gained during the internship contributed to the motivation and formulation in this paper. We also thank Dr Joseph Wang for helpful discussions.

\bibliography{cost_sensitive_bib}
\bibliographystyle{plain}
\clearpage
\section{Appendix}
\subsection{\textsc{DynaMod-Lstsq}}
\paragraph{Symetrized distance:} Besides the KL-divergence used in Section 3.0.1, we can also use a squared distance of $g$ to the log likelihood ratio of $q$:
$D(q(\cdot|x),p_\theta(\cdot|x))=(\log\frac{q(0|x)}{q(1|x)}-\log\frac{\Pr{(z=0|x;g)}}{\Pr(z=1|x;g)})^2=(\log\frac{q(0|x)}{q(1|x)}-g(x))^2$.
We found this distance function to perform better than KL-divergence in the local-remote setup. We still follow an alternating minimization approach. To keep the presentation simply, we assume $g, f_1$ to be linear classifiers and there is no feature costs involved.
To minimize over $q$, we must solve 
\begin{equation}
\begin{array}{rlll}\tag{OPT5}\label{eq:OPT5}
\displaystyle \min_{q_i\in [0,1]} &  \multicolumn{2}{l}{\frac{1}{N} \sum_{i=1}^{N} \left [(1-q_i)A_i+(\log\frac{q_i}{1-q_i}-g(x^{(i)}))^2\right ]} \\
\textrm{s.t.} &  \frac{1}{N} \sum_{i=1}^{N} q_i \leq \text{P}_{\text{full}}, \end{array}
\end{equation}
where $q_i=q(z=0|x^{(i)})$, $A_i=\log(1+e^{-y^{(i)}f_1^Tx^{(i)}})+\log p(y^{(i)}|z^{(i)}=1;f_0)$. Unlike (OPT3), this optimization problem no longer has a closed-form solution. Fortunately, the $q_i$'s in the objective are decoupled and there is only one coupling constraint. We can solve this problem using an ADMM approach \cite{Boyd:2011:DOS:2185815.2185816}.
To optimize over $g$, we simply need to solve a linear least squares problem:
\begin{equation}\tag{OPT6}\label{eq:OPT6}
\displaystyle \min_{g} \frac{1}{N} \sum_{i=1}^{N} (\log\frac{q_i}{1-q_i}-g^T(x^{(i)}))^2.
\end{equation}
To optimize over $f_1$, we solve a weighted logistic regression problem:
\begin{equation}\tag{OPT7}\label{eq:OPT7}
\displaystyle \min_{f_1} \frac{1}{N} \sum_{i=1}^{N} (1-q_i)\log(1+e^{-y^{(i)}f_1^Tx^{(i)}}).
\end{equation}
We shall call the above algorithm \textsc{DynaMod-Lstsq}, summarized in Algorithm~\ref{alg:dynamod-lstsq}.
\begin{algorithm}[tb]
	\caption{\textsc{DynaMod-Lstsq}}
	\label{alg:dynamod-lstsq}
	\begin{algorithmic}
		\STATE {\bfseries Input:} $(x^{(i)},y^{(i)}),B$
		\STATE Train a full accuracy model $f_0$.
		\STATE Initialize $g, f_1$.
		\REPEAT
		\STATE Solve (OPT5) for $q$ given $g, f_1$. 
		\STATE Solve (OPT6) for $g$ given $q$.
		\STATE Solve (OPT7)for $f_1$ given $q$.
		\UNTIL{convergence}
	\end{algorithmic}
\end{algorithm}

\subsection{Experimental Details}
We provide detailed parameter settings and steps for our experiments here.
\subsection{Synthetic Experiment}
We generate 4 clusters on a 2D plane with centers: (1,1), (-1,1), (-1,-1), (-1, -3) and Gaussian noise with standard deviation of 0.01. The first two clusters have 20 examples each and the last two clusters have 15 examples each. We sweep the regularization parameter of L1-regularized logistic regression and recover feature 1 as the sparse subset, which leads to sub-optimal adaptive system.
On the other hand, we can easily train a RBF SVM classifier to correctly classify all clusters and we use it as $f_0$. If we initialize $g$ and $f_1$ with Gaussian distribution centered around 0, \textsc{DynaMod-Lin} with can often recover feature 2 as the sparse subset and learn the correct $g$ and $f_1$. Or, we could initialize $g=(1,1)$ and $f_1=(1,1)$ then \textsc{DynaMod-Lin} can recover the optimal solution.

\subsection{Letters Dataset \cite{UCI_repository}}
 This letters recognition dataset contains 20000 examples with 16 features, each of which is assigned unit cost. We binarized the labels so that the letters before "N" is class 0 and the letters after and including "N" are class 1.
 We split the examples 12000/4000/4000 for training/validation/test sets. 
 We train RBF SVM and RF (500 trees) with cross-validation as $f_0$. RBF SVM achieves the higher accuracy of 0.978 compared to RF 0.961.
 
 To run the greedy algorithm, we first cross validate L1-regularized logistic regression with 20 C parameters in logspace of [1e-3,1e1]. For each C value, we obtain a classifier and we order the absolute values of its components and threshold them at different levels to recover all 16 possible supports (ranging from 1 feature to all 16 features). We save all such possible supports as we sweep C value. Then for each of the supports we have saved, we train a L2-regularized logistic regression only based on the support features with regularization set to 1 as $f_1$. The gating $g$ is then learned using L2-regularized logistic regression based on the same feature support and pseudo labels of $f_1$ - 1 if it is correctly classified and 0 otherwise. To get different cost-accuracy tradeoff, we sweep the class weights between 0 and 1 so as to influence $g$ to send different fractions of examples to the $f_0$. 
 
 To run \textsc{DynaMod-Lin}, we initialize $g$ to be 0 and $f_1$ to be the output of the L2-regularized logistic regression based on all the features. We then perform the alternative minimization for 50 iterations and sweep $\gamma$ between [1e-4,1e0] for 20 points and $\text{P}_\text{full}$ in [0.1,0.9] for 9 points. 
 
 To run \textsc{DynaMod-Gbrt}, we use 500 depth 4 trees for $g$ and $f_1$ each. We
 initialize $g$ to be 0 and $f_1$ to be the GreedyMiser output of 500 trees. We then perform the alternative minimization for 30 iterations and sweep $\gamma$ between [1e-1,1e2] for 10 points in logspace and $\text{P}_\text{full}$ in [0.1,0.9] for 9 points. In addition, we also sweep the learning rate for GBRT for 9 points between [0.1,1].
 
 For fair comparison, we run \textsc{GreedyMiser} with 1000 depth 4 trees so that the model size matches that of \textsc{DynaMod-Gbrt}. The learning rate is swept between [1e-5,1] with 20 points and the $\lambda$ is swept between [0.1, 100] with 20 points.
 
 Finally, we evaluate all the resulting systems from the parameter sweeps of all the algorithms on validation data and choose the efficient frontier and use the corresponding settings to evaluate and plot the test performance. 
 
 \subsection{MiniBooNE Particle Identification and Forest Covertype Datasets \cite{UCI_repository}:}
 The MiniBooNE data set is a binary classification task to distinguish electron neutrinos from muon neutrinos. There are $45523/19510/65031$ examples in training/validation/test sets. Each example has 50 features, each with unit cost. 
 The Forest data set contains cartographic variables to predict 7 forest cover types. There are $36603/15688/58101$ examples in training/validation/test sets. Each example has 54 features, each with unit cost.

 We use the unpruned RF of \textsc{BudgetPrune} \cite{NanNIPS2016} as $f_0$ (40 trees for both datasets.) The settings for \textsc{DynaMod-Gbrt} are the following.
 For MiniBooNE we use 100 depth 4 trees for $g$ and $f_1$ each. We
  initialize $g$ to be 0 and $f_1$ to be the GreedyMiser output of 100 trees. We then perform the alternative minimization for 50 iterations and sweep $\gamma$ between [1e-1,1e2] for 20 points in logspace and $\text{P}_\text{full}$ in [0.1,0.9] for 9 points. In addition, we also sweep the learning rate for GBRT for 9 points between [0.1,1].
  For Forest we use 500 depth 4 trees for $g$ and $f_1$ each. We
    initialize $g$ to be 0 and $f_1$ to be the GreedyMiser output of 500 trees. We then perform the alternative minimization for 50 iterations and sweep $\gamma$ between [1e-1,1e2] for 20 points in logspace and $\text{P}_\text{full}$ in [0.1,0.9] for 9 points. In addition, we also sweep the learning rate for GBRT for 9 points between [0.1,1].
    
For fair comparison, we run \textsc{GreedyMiser} with 200 depth 4 trees so that the model size matches that of \textsc{DynaMod-Gbrt} for MiniBooNE. We run \textsc{GreedyMiser} with 1000 depth 4 trees so that the model size matches that of \textsc{DynaMod-Gbrt} for Forest.

   Finally, we evaluate all the resulting systems from the parameter sweeps on validation data and choose the efficient frontier and use the corresponding settings to evaluate and plot the test performance. 
 
 \subsection{Yahoo! Learning to Rank\cite{YahooChallenge2010}:}
  This ranking dataset consists of 473134 web documents and 19944 queries. Each example is associated with features of a query-document pair together with the relevance rank of the document to the query. There are 519 such features in total; each is associated with an acquisition cost in the set \{1,5,20,50,100,150,200\}, which represents the units of CPU time required to extract the feature and is provided by a Yahoo! employee. The labels are binarized into relevant or not relevant. The task is to learn a model that takes a new query and its associated documents and produce a relevance ranking so that the relevant documents come on top, and to do this using as little feature cost as possible. The performance metric is Average Precision @ 5 following \cite{NanNIPS2016}. 

   We use the unpruned RF of \textsc{BudgetPrune} \cite{NanNIPS2016} as $f_0$ (140 trees for both datasets.) The settings for \textsc{DynaMod-Gbrt} are the following. we use 100 depth 4 trees for $g$ and $f_1$ each. We
       initialize $g$ to be 0 and $f_1$ to be the \textsc{GreedyMiser} output of 100 trees. We then perform the alternative minimization for 20 iterations and sweep $\gamma$ between [1e-1,1e3] for 30 points in logspace and $\text{P}_\text{full}$ in [0.1,0.9] for 9 points. In addition, we also sweep the learning rate for GBRT for 9 points between [0.1,1].

For fair comparison, we run \textsc{GreedyMiser} with 200 depth 4 trees so that the model size matches that of \textsc{DynaMod-Gbrt} for Yahoo. 
  
  Finally, we evaluate all the resulting systems from the parameter sweeps on validation data and choose the efficient frontier and use the corresponding settings to evaluate and plot the test performance. 
  
  \subsection{Local Constrained Systems:}

  In our experiments, we perform classification on real datasets using gradient boosted regression trees. For a fixed depth, the complexity of a model is measured by the number of trees it uses. On the Cifar10\cite{CIFAR10}, MiniBooNE and Forest datasets, we train 500 trees of depth 5 with various learning rates from 0.1 to 1 and choose the best one using validation data as the complex model. Now assume a device can only host a local model of 10 trees. Our goal is to train a small model together with a gating function such that the overall accuracy of the local - remote system is maintained at the same level of the complex model and send as few examples to the cloud as possible. 
  
  Before describing our dynamic model selection algorithm for this problem, it is helpful to first consider a simple confidence-based approach and binary classification task. This approach finds the best depth 5 GBRT of 10 trees using validation data as the small model, denoted as $f_1$. It also determines a threshold $\tau$ of the absolute margin using the training/validation data such that a desired fraction of examples are sent to the complex model. During prediction, the example $x$ is run through the small model $\mathbf{f_1}$. If $|f_1(x)|<\tau$ then $x$ is sent to the complex classifier; otherwise it is classified by the local model $\text{sgn}(f_1(x))$.
 Despite its simplicity, this approach often performs quite well. One of the reasons is that the absolute value of the margin $|\mathbf{h}(x)|$ can be viewed as confidence in classification. Thus, the policy intuitively classifies an  example locally if it has high confidence and sends the example to the complex model if it has low confidence. Note the gating function here is in a sense integrated with the local model $\mathbf{h}$ - it does not require more evaluating more trees. 
  Next, we consider how our approach can be used to improve upon the confidence-based approach. Given the local GBRT model $f_1$, we can view it as a linear classifier in the transformed feature space: $x \to \phi(x)\in \Re^d$, where $d$ is the total number of leaves in the ensemble of trees and each element of $\phi(x)$ corresponds to a leaf. $\phi_j(x)=1$ if $x$ ends up in the $j$th leaf and $\phi_j(x)=0$ otherwise. Clearly, $f_1(x) = w^T\phi(x)$, where $w$ is a vector of leaf weights. To mimic the confidence-based policy, we augment $\phi(x)$ by the confidence $|f_1(x)|$ such as $\hat{\phi}(x) = [\phi(x) ; |f_1(x)|] \in \Re^{d+1}$. The gating function is based on feature $\hat{\phi}$: $g(x)=g^T\hat{\phi}+g_0$. To simplify notation, we will use $\phi$ instead of $\hat{\phi}$ to denote the augmented feature transform. Similar to the confidence approach, the gating function incurs little overhead as it reuses the same tree structure of the local model and only learn a different set of leaf weights. During prediction this additional overhead is negligible compared to storing and evaluating the local tree structure. 
  Our approach involves learning the weights of the local model $f_1$ and the gating function $g$ iteratively. We begin by initializing $f_1$ using the original leaf weights; we initialize $g$ to be 0's except -1 for the last element that corresponds to the absolute margin feature and the offset $g_0 = \tau$. Clearly, this initialization corresponds exactly to the myopic approach and we start from here.   We found that \textsc{DynaMod-Lstsq} gives better results than \textsc{DynaMod-Lin}. One possible reason is that the optimization over $g$ in \textsc{DynaMod-Lin} (can be viewed as an M-projection) becomes difficult as the objective becomes too flat for a linear parameterization.

\end{document}